\pdfoutput=1
\PassOptionsToPackage{table,xcdraw}{xcolor}
\documentclass{l4dc2027}
\makeatletter
\jmlrvolume{}\jmlrpages{}\jmlryear{}
\renewcommand*{\@jmlrproceedings}{Preprint. Under review.}
\renewcommand*{\@titlefoot}{\scriptsize\copyright\space 2026
  \@jmlr@authors.\hfill}
\makeatother
\usepackage{times}
\usepackage{soul} %

\newcommand{\corlPhysicsHz}{250}
\newcommand{\corlControlTickMs}{40}
\newcommand{\corlMpcHz}{50}
\newcommand{\corlSwingDurationMin}{0.1}
\newcommand{\corlSwingDurationMax}{0.3}
\newcommand{\corlMaxConcurrentSwingLegs}{2}

\newcommand{\corlScanResCm}{1}
\newcommand{\corlGridReachCm}{18}
\newcommand{\corlMarginCm}{2}
\newcommand{\corlNumLegs}{4}
\newcommand{\corlTrainCandidatesPerLeg}{16}
\newcommand{\corlTotalCandidates}{65}
\newcommand{\corlNomLiftS}{0.02}       %
\newcommand{\corlNomSwingS}{0.18}      %
\newcommand{\corlNomHorizonS}{0.29}    %
\newcommand{\corlAnnealSteps}{4}
\newcommand{\corlAnnealTauStartCm}{2}  %
\newcommand{\corlAnnealTauEndCm}{0.5}  %

\newcommand{\corlHistTicks}{25}
\newcommand{\corlHistSeconds}{1}

\newcommand{\corlMixerLayers}{2}
\newcommand{\corlMixerDim}{128}
\newcommand{\corlMixerHeads}{4}
\newcommand{\corlMixerFF}{256}
\newcommand{\corlParamsPerCandidateK}{99}   %
\newcommand{\corlParamsGaitNetK}{464} %
\newcommand{\corlDurationExploreStd}{0.01}

\newcommand{\corlMpcHorizon}{0.2}
\newcommand{\corlOvershootSlowMin}{1.3}
\newcommand{\corlOvershootSlowMax}{1.9}
\newcommand{\corlOvershootFastMin}{2.9}
\newcommand{\corlOvershootFastMax}{3.9}

\newcommand{\corlIdmHidden}{256}
\newcommand{\corlIdmFootsteps}{180{,}681}
\newcommand{\corlIdmHeldOutPct}{10}
\newcommand{\corlIdmRmsCm}{0.89}
\newcommand{\corlIdmIdentityRmsCm}{2.61}

\newcommand{\corlParallelEnvs}{1024}
\newcommand{\corlRolloutSteps}{100}
\newcommand{\corlPpoEpochs}{8}
\newcommand{\corlMiniBatches}{8}
\newcommand{\corlLearningRate}{3 \times 10^{-4}}
\newcommand{\corlClip}{0.3}
\newcommand{\corlEntropyCoef}{0.02}
\newcommand{\corlGamma}{0.99}
\newcommand{\corlLambda}{0.95}
\newcommand{\corlEpisodeSeconds}{20}
\newcommand{\corlAliveWeight}{0.4}
\newcommand{\corlTrackXyStd}{0.1}
\newcommand{\corlTrackYawStd}{0.2}
\newcommand{\corlActionPenalty}{0.4}      %
\newcommand{\corlTerminationPenalty}{200} %
\newcommand{\corlTiltLimitDeg}{20}
\newcommand{\corlMinBaseHeight}{0.15}
\newcommand{\corlFootDepthLimit}{0.05}
\newcommand{\corlCmdVxMax}{0.2}
\newcommand{\corlCmdYawMax}{0.2}
\newcommand{\corlTerrainCellCm}{12.5}
\newcommand{\corlVoidDepth}{0.5}
\newcommand{\corlNumLevels}{12}
\newcommand{\corlMaxVoidPct}{40}
\newcommand{\corlCurriculumProgress}{0.3}
\newcommand{\corlWalkPUp}{0.1}
\newcommand{\corlWalkPDown}{0.3}
\newcommand{\corlWalkRandomPct}{25}
\newcommand{\corlRandTrackMin}{0.3}
\newcommand{\corlRandPlanarMin}{0.5}
\newcommand{\corlRandApexMin}{0.07}
\newcommand{\corlRandApexMax}{0.12}
\newcommand{\corlRandBiasCm}{2}
\newcommand{\corlRandNoiseCm}{1}
\newcommand{\corlTrainHours}{4}

\newcommand{\corlSimSpeedSlow}{0.1}
\newcommand{\corlSimSpeedFast}{0.2}

\newcommand{\corlOcsTrialsPerPoint}{80}
\newcommand{\corlOcsCheckpoint}{300}  %
\newcommand{\corlNumSeedsGaitNet}{five}
\newcommand{\corlOcsTrackM}{3}
\newcommand{\corlOcsTimeFactor}{1.5}
\newcommand{\corlOcsLandRmsNoIdm}{6}
\newcommand{\corlOcsLandRmsIdm}{2}
\newcommand{\corlOcsMaxVoidPct}{70}
\newcommand{\corlOcsSpeedMin}{0.05}
\newcommand{\corlOcsSpeedMax}{0.25}

\newcommand{\corlBenchTileCm}{12.5}
\newcommand{\corlBenchVoidLow}{26}
\newcommand{\corlBenchVoidMid}{38}
\newcommand{\corlBenchVoidHigh}{49}
\newcommand{\corlBenchIsaacVoidLow}{20}
\newcommand{\corlBenchIsaacVoidMid}{30}
\newcommand{\corlBenchIsaacVoidHigh}{40}
\newcommand{\corlBenchSpeedSlow}{0.10}
\newcommand{\corlBenchSpeedFast}{0.20}
\newcommand{\corlBenchTrialSeconds}{20}
\newcommand{\corlBenchTrialsPerCell}{100}
\newcommand{\corlBenchTrialsPerArm}{600}
\newcommand{\corlBenchFull}{85.1}  %
\newcommand{\corlBenchTrot}{67.8}  %
\newcommand{\corlBenchFullDense}{72.9}  %
\newcommand{\corlBenchTrotDense}{43.0}  %
\newcommand{\corlBenchDiag}{82.9}  %
\newcommand{\corlBenchNoIdm}{17.1}  %
\newcommand{\corlBenchDiagDense}{69.6}  %
\newcommand{\corlBenchNoIdmDense}{13.2}  %
\newcommand{\corlSweepDiag}{73.2}
\newcommand{\corlBenchSingle}{39.6}  %
\newcommand{\corlNumTrainSeeds}{five}
\newcommand{\corlAuthorityCm}{10}
\newcommand{\corlBenchTrainIfm}{5.5}  %
\newcommand{\corlBenchTrainIfmDense}{4.3}  %
\newcommand{\corlBenchOffset}{60.0}  %
\newcommand{\corlBenchOffsetDense}{50.1}  %
\newcommand{\corlBenchStaticWalk}{65.5}  %
\newcommand{\corlBenchStaticWalkDense}{40.0}  %
\newcommand{\corlWalkPeriod}{1.2}
\newcommand{\corlBenchMlpMean}{71.3}  %
\newcommand{\corlTrotSwing}{0.3}
\newcommand{\corlTrotPeriod}{0.6}

\newcommand{\corlSweepFull}{82.6}          %
\newcommand{\corlSweepTrot}{64.1}          %

\newcommand{\corlGaitTrialsGaitNet}{549}   %
\newcommand{\corlGaitTrialsTrot}{504}
\newcommand{\corlGaitExampleSeed}{134} %
\newcommand{\corlGaitOverlapPairs}{24{,}793}  %
\newcommand{\corlGaitDiagPairPct}{98}      %
\newcommand{\corlGaitStepsSlow}{5.5}       %
\newcommand{\corlGaitStepsFast}{7.2}       %
\newcommand{\corlGaitStepsTrot}{6.7}       %
\newcommand{\corlGaitStrideCv}{0.61}       %
\newcommand{\corlGaitStrideCvTrot}{0.02}   %
\newcommand{\corlGaitTrotPeriodMeasured}{0.598} %
\newcommand{\corlGaitTrotSwingMeasured}{0.235}  %
\newcommand{\corlGaitLiftLagMs}{42}        %

\newcommand{\corlSeedsGaitNet}{5}  %
\newcommand{\corlBenchFullMin}{78.3}  %
\newcommand{\corlBenchFullMax}{89.2}  %
\newcommand{\corlBenchDiagMin}{75.7}  %
\newcommand{\corlBenchDiagMax}{86.3}  %
\newcommand{\corlBenchSingleMin}{37.2}  %
\newcommand{\corlBenchSingleMax}{42.8}  %
\newcommand{\corlBenchMlpSeedMin}{68.7}  %
\newcommand{\corlBenchMlpSeedMax}{73.3}  %
\newcommand{\corlBenchNoIdmMin}{16.3}  %
\newcommand{\corlBenchNoIdmMax}{18.2}  %
\newcommand{\corlBenchTrainIfmMin}{3.8}  %
\newcommand{\corlBenchTrainIfmMax}{7.2}  %
\newcommand{\corlBenchOffsetMin}{57.5}  %
\newcommand{\corlBenchOffsetMax}{63.0}  %
\newcommand{\corlBenchSingleDense}{35.6}  %
\newcommand{\corlBenchMlpDense}{61.5}  %

\newcommand{\corlGaitEventWindowMs}{66}      %
\newcommand{\corlGaitEventRepeatPct}{15.9}   %
\newcommand{\corlGaitEventRepeatPctTrot}{0}
\newcommand{\corlGaitLegTwicePct}{6.1}       %
\newcommand{\corlGaitLegTwicePctTrot}{0}
\newcommand{\corlGaitFourAllPct}{40}         %
\newcommand{\corlGaitFourAllPctTrot}{99}     %
\newcommand{\corlGaitSwingMedian}{0.128}     %
\newcommand{\corlGaitSwingMedianTrot}{\corlGaitTrotSwingMeasured} %
\newcommand{\corlGaitSwingsGaitNet}{55{,}170} %
\newcommand{\corlGaitSwingsTrot}{53{,}431}    %

\usepackage{amsfonts}
\usepackage{textcomp}
\graphicspath{{./}}
\DeclareGraphicsExtensions{.pdf,.png,.jpg}
\definecolor{linkgreen}{HTML}{1E7B4F}
\hypersetup{colorlinks=true,linkcolor=linkgreen,citecolor=linkgreen,
  urlcolor=linkgreen}
\usepackage{tikz}
\usetikzlibrary{arrows.meta,calc,positioning}
\usepackage{fontawesome5}
\definecolor{ink}{HTML}{1F2937}       %
\definecolor{muted}{HTML}{6B7280}     %
\definecolor{rgA}{HTML}{F5F6F8}\definecolor{rgAd}{HTML}{C3C9D2}   %
\definecolor{rgB}{HTML}{EEF1FD}\definecolor{rgBd}{HTML}{A9B4F4}   %
\definecolor{rgC}{HTML}{EBF7F1}\definecolor{rgCd}{HTML}{8FD3B3}   %
\definecolor{acB}{HTML}{4F46E5}       %
\definecolor{acC}{HTML}{0F9F6E}       %
\definecolor{acT}{HTML}{D97706}       %
\definecolor{acF}{HTML}{2B8CC4}       %
\definecolor{tkS}{HTML}{E0E7FF}       %
\definecolor{tkC}{HTML}{FEF3C7}\definecolor{tkCd}{HTML}{F2B43A} %
\definecolor{tkN}{HTML}{F3F4F6}       %
\definecolor{rgT}{HTML}{FFF8EE}\definecolor{rgTd}{HTML}{F1C48A}   %
\definecolor{rgS}{HTML}{EEF1FD}\definecolor{rgSd}{HTML}{A9B4F4}   %
\definecolor{rgG}{HTML}{EBF7F1}\definecolor{rgGd}{HTML}{8FD3B3}   %
\tikzset{
  arr/.style={->, line width=0.7pt, draw=ink},
  tarr/.style={->, line width=0.8pt, draw=acT, dash pattern=on 3pt off 2pt},
  region/.style 2 args={fill=#1, draw=#2, line width=0.7pt,
    dash pattern=on 3pt off 2pt, rounded corners=6pt},
  rlabel/.style={font=\sffamily\footnotesize, text=muted, anchor=south west,
    inner sep=0pt},
  sub/.style={font=\sffamily\scriptsize, text=muted, align=center, inner sep=1pt},
  tok/.style 2 args={fill=#1, draw=#2, line width=0.5pt, rounded corners=1.5pt,
    minimum width=7.5mm, minimum height=5.2mm, inner sep=1pt},
  mod/.style={fill=white, draw=#1, line width=0.7pt, rounded corners=4pt,
    align=center, inner sep=2.5pt, minimum height=7mm},
  pics/attn/.style={code={
    \foreach \i in {0,...,4}{\foreach \j in {0,...,4}{
      \draw[#1!55, line width=0.25pt] (\i*0.26-0.52,0.24) -- (\j*0.26-0.52,-0.24);}}
    \foreach \i in {0,...,4}{
      \fill[#1] (\i*0.26-0.52,0.24) circle (0.042);
      \fill[#1] (\i*0.26-0.52,-0.24) circle (0.042);}
  }},
  pics/mlp/.style={code={
    \foreach \i in {0,...,2}{\foreach \j in {0,...,3}{
      \draw[#1!55, line width=0.25pt] (-0.42,\i*0.24-0.24) -- (0,\j*0.24-0.36);}}
    \foreach \i in {0,...,3}{\foreach \j in {0,...,1}{
      \draw[#1!55, line width=0.25pt] (0,\i*0.24-0.36) -- (0.42,\j*0.24-0.12);}}
    \foreach \i in {0,...,2}{\fill[#1] (-0.42,\i*0.24-0.24) circle (0.042);}
    \foreach \i in {0,...,3}{\fill[#1] (0,\i*0.24-0.36) circle (0.042);}
    \foreach \i in {0,...,1}{\fill[#1] (0.42,\i*0.24-0.12) circle (0.042);}
  }},
  gfig/.style={font=\sffamily\small, text=ink, >={Stealth[length=2.1mm,width=1.6mm]}},
  ptitle/.style={font=\sffamily\footnotesize\bfseries, text=ink, anchor=north west, inner sep=0pt},
  band/.style 2 args={fill=#1, draw=#2, line width=0.6pt, rounded corners=5pt},
  alab/.style={font=\footnotesize, text=ink, inner sep=2pt},   %
  tlab/.style={font=\sffamily\scriptsize, inner sep=2pt},     %
}

\usepackage{multirow}
\usepackage{booktabs}
\usepackage{fmtcount}
\usepackage{array}
\usepackage{float}
\usepackage{dblfloatfix}
\makeatletter
\@ifundefined{proposition}{\newtheorem{proposition}{Proposition}}{}
\@ifundefined{lemma}{}{}
\@ifundefined{remark}{}{}
\@ifundefined{assumption}{\newtheorem{assumption}{Assumption}}{}
\makeatother
\newenvironment{idmproof}[1]%
  {\par\medskip\noindent\textbf{Proof of #1.}\ }%
  {\hfill$\square$\par\medskip}

\definecolor{tabink}{HTML}{111111}
\definecolor{tabmuted}{HTML}{454C55}
\definecolor{tabaccent}{HTML}{1A5FB4}
\newcommand{\thd}[1]{\textcolor{tabink}{\bfseries #1}}      %

\title[CADENCE]{CADENCE: Learning Acyclic Gaits for Model-Based
  Quadruped Locomotion}

\author{%
 \Name{Owen Sullivan\nametag{\thanks{Equal contribution.}}} \Email{opsullivan85@gmail.com}\\
 \addr Worcester Polytechnic Institute
 \AND
 \Name{Sourav Selvaraj\nametag{\footnotemark[1]}} \Email{sselvaraj@wpi.edu}\\
 \addr Worcester Polytechnic Institute
 \AND
 \Name{Mahdi Agheli\nametag{\thanks{Corresponding author.}}} \Email{mmaghelih@wpi.edu}\\
 \addr Worcester Polytechnic Institute%
}

\begin{document}

\maketitle

\makeatletter\newcommand{\teasercaptype}{\def\@captype{figure}}\makeatother
\vspace{-1.4em}
\begin{center}
\teasercaptype
\includegraphics[width=\textwidth]{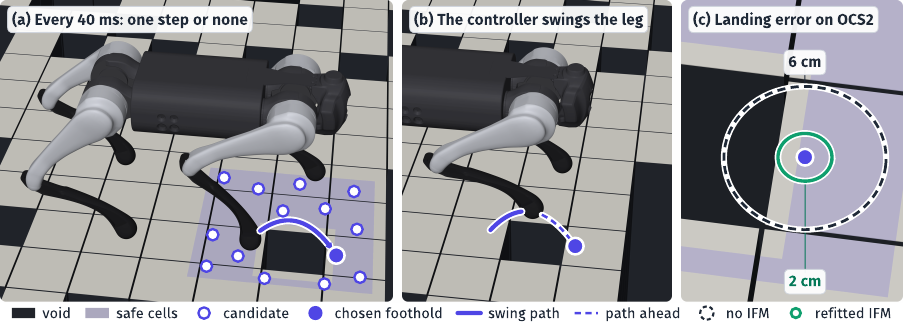}
\vspace{-1.4em}
\caption{CADENCE decides when and where a model-based controller steps,
  shown here on a posed MuJoCo scene with an illustrative step.
  (a)~Every \corlControlTickMs\,ms, it either waits or starts one step by
  picking a leg and one of that leg's safe footholds, at most
  \corlTotalCandidates{} choices in all, and it sets the swing time.
  (b)~The controller then swings the leg to the chosen foothold.
  (c)~On OCS2, refitting the inverse footstep model (IFM) brings the RMS
  landing error on flat ground down to about the candidate margin.}
\label{fig:teaser}
\end{center}
\vspace{-0.8em}

\begin{abstract}%
  Model-based quadruped controllers usually step on a cyclic gait, which
  lifts the legs in a fixed order on a fixed period, wherever the safe
  footholds are. We present CADENCE, a learned step planner with a small
  action space. Every \corlControlTickMs\,ms, it picks one of at most
  \corlTotalCandidates{} discrete choices: no step, or one leg's step to
  a safe foothold. It also predicts a swing time.
  We train CADENCE by reinforcement learning on top of a convex model
  predictive controller (MPC). Nothing prescribes a gait: there is no
  gait schedule, and the reward has no gait term, only a cost per step.
  The one structural constraint is that at most
  \numberstringnum{\corlMaxConcurrentSwingLegs} legs may swing at once.
  The policy we analyzed developed an acyclic gait, with no fixed period
  or stepping order, yet it still paired diagonal legs. An
  inverse footstep model (IFM), fitted on a controller's logged steps,
  turns each foothold into a command. With only the IFM refitted, on
  fewer than 20{,}000 steps, CADENCE ran frozen on OCS2's
  nonlinear MPC. In simulated trials on a range of terrains (26\% to 49\% voids) and speeds (0.1\,m/s and 0.2\,m/s), it succeeded
  in \corlBenchFull\% (mean of \corlNumSeedsGaitNet{} seeds),
  against \corlBenchTrot\% for OCS2-Trot,
  \corlBenchStaticWalk\% for OCS2-Single, and \corlBenchSingle\% in a static walk ablation of CADENCE. At \corlBenchVoidHigh\% voids, it
  succeeded in \corlBenchFullDense\%, against \corlBenchTrotDense\% for
  OCS2-Trot. %

\end{abstract}


\section{Introduction}
\label{sec:introduction}

\looseness=-1 To cross level ground with holes, or \emph{voids}
(Fig.~\ref{fig:teaser}), a quadruped must decide which leg to lift,
where to place it, when and for how long to swing. Model-based controllers
\citep{dicarlo_dynamic_2018, ocs2} track such steps well, but they take
them from a cyclic gait: a trot lifts diagonal leg pairs and a static
walk lifts one leg at a time, in a fixed order and on a fixed period,
wherever the safe footholds are. On sparse footholds, a cycle can
therefore call for a step when no safe foothold is in reach.

\looseness=-1 Choosing the gait online instead raises two problems.
First, contact decisions are discrete, so choosing them inside a model
predictive controller (MPC) makes its optimization mixed-integer and increases the search space and compute required to find solutions
\citep{Wensing2022Nov}. Second, a planner outside the controller only
commands where a foot should land, and the controller lands the foot at
an offset from that command. This offset differs between controllers and
can exceed the margin that keeps a chosen foothold clear of voids, so a
planner trained with one controller can miss safe footholds with
another.

\looseness=-1 Gait-free planners choose contact timing by kinematic
rules or tree search \citep{bjelonic_whole-body_2021,
amatucci_monte_2022, taouil_non-gaited_2024}, or learn it for a given
MPC without choosing footholds \citep{patrizi_rl-augmented_2026}.
Hybrid learned approaches choose the parameters of a cyclic gait, per-foot contact
states, or footholds under a preset timing \citep{yang_fast_2021,
margolis_learning_2021, gangapurwala_rloc_2022,
bratta_contactnet_2024}. At the footstep level,
\citet{akizhanov_learning_2024} correct a nonlinear MPC's contact
targets by hindsight regression, but within a fixed gait. To our
knowledge, no learned gait-free footstep planner has moved between
model-based controllers.

\begin{figure*}[t]
\centering
\includegraphics[width=0.99\textwidth]{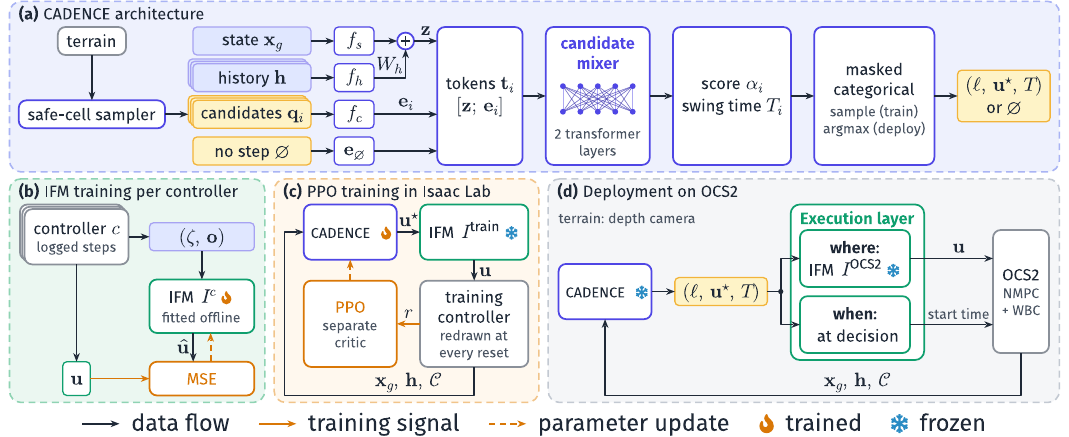}
\vspace{-0.5em}
\caption{CADENCE overview. (a)~Given the state $\mathbf x_g$ and history
  $\mathbf h$, a candidate mixer scores the safe candidates
  $\mathbf q_i \in \mathcal C$ and the no-step option $\varnothing$.
  (b)~Each controller $c$ gets its own IFM $I^c$, fitted offline on its
  logged steps. (c)~PPO keeps the training controller's IFM frozen.
  (d)~On OCS2, CADENCE stays frozen and only the IFM changes.}
\vspace{-1.4em}
\label{fig:overview}
\end{figure*}

This work presents CADENCE (Fig.~\ref{fig:overview}), which builds on the
foundation previously developed through the first author's
thesis work, GaitNet \citep{Sullivan2025}. CADENCE is a learned planner with a small
action space which works with different model-based controllers. Every
\corlControlTickMs\,ms, a \emph{tick}, it makes one of at most
\corlTotalCandidates{} discrete choices: the \emph{no-step option}, or
one leg's step to one of \corlTrainCandidatesPerLeg{} safe cells near
its hip, the \emph{candidates} (Fig.~\ref{fig:action}), plus a swing
time. The swing timing, order, overlap, and pairings emerge from these
choices, with no gait schedule, phase variable, or gait reward. CADENCE
scores each candidate on its own. A small transformer, the \emph{candidate mixer},
compares the options with each
other, a history adds the recent motion, and an \emph{inverse footstep
model} (IFM), fitted by hindsight regression on logged steps, turns each
chosen foothold into a command accounting for controller error. Trained
with proximal policy optimization (PPO) on the convex MPC, CADENCE runs
frozen on OCS2's nonlinear MPC (NMPC) and whole-body controller (WBC),
with only the IFM refitted on fewer than 20{,}000 OCS2 steps. We
evaluate both CADENCE and its ablations against OCS2's own cyclic gaits.
Our contributions are:

\begin{enumerate}\setlength{\itemsep}{0pt}
  \item A learned step planner for model-based controllers.
    Its network compares the choices of a small per-tick action space
    with a candidate mixer and a history
    (Secs.~\ref{subsec:method-step}--\ref{subsec:method-network}).
  \item An emergent acyclic gait, measured and compared with a
    model-based controller's own cyclic gaits: on OCS2, CADENCE's gait
    had neither a fixed period nor a fixed order, yet mostly paired
    diagonal legs, and CADENCE succeeded more often than OCS2-Trot and
    OCS2-Single (Sec.~\ref{subsec:exp-transfer}; acyclicity measured on
    one earlier run of one policy).
  \item Transfer between model-based controllers by refitting only the
    IFM (Secs.~\ref{subsec:method-idm}
    and~\ref{subsec:method-deployment}), where simpler corrections fell
    short (Sec.~\ref{subsec:exp-why}). The refit uses steps logged on
    the new controller, so the transfer is not data-free.
\end{enumerate}

\section{Related Work}
\label{sec:related}
\vspace{-2em}
\noindent\textbf{Gaits in model-based control.} \looseness=-1
Model-based controllers usually take their contact timing from a cyclic
gait \citep{dicarlo_dynamic_2018, grandia_perceptive_2022}, while
planners optimize timing or footholds \citep{winkler_gait_2018,
jenelten_tamols_2022}. Gait-free planners choose contacts online by
kinematic rules \citep{bjelonic_whole-body_2021} or tree search
\citep{amatucci_monte_2022, taouil_non-gaited_2024}, or let a learned
agent set an MPC's contact timing, from which acyclic gaits also emerge
\citep{patrizi_rl-augmented_2026}. However, the tree searches rate
contact sequences by the controller's own optimal cost, and the learned
agent chooses no footholds and is trained for the MPC it runs with,
whereas CADENCE sends an unchanged controller one step at a time,
foothold included, and moves between controllers.

\noindent\textbf{Learned gaits and footholds.} \looseness=-1 End-to-end
policies decide contacts implicitly, in a large action space of joint
or foot targets \citep{lee_learning_2020, zhang_learning_2023}. Hybrids
learn the parameters of a cyclic gait or per-foot contact states with
heuristic footholds \citep{yang_fast_2021, margolis_learning_2021,
da_learning_2020, yang_cajun_2023}, or footholds under a fixed gait
\citep{gangapurwala_rloc_2022, villarreal_fast_2019}. Close to CADENCE
in its action space is ContactNet \citep{bratta_contactnet_2024}, a
supervised network that ranks discrete footholds near each hip and so
chooses which legs step and where, in an acyclic order. However, its
walk or trot setting presets the swing times and how many legs swing,
and it replans only after each touchdown.

\noindent\textbf{Transfer between controllers.} \looseness=-1 Policies
move to a new system through an inverse dynamics model fitted on its
data \citep{christiano_transfer_2016, xie_fada_2026}, a grounded
simulator \citep{hanna_grounded_2017}, or online adaptation
\citep{kumar_rma_2021}. \citet{akizhanov_learning_2024} correct an
NMPC's contact targets with a network fitted on the contacts it reached,
within a fixed gait and on one controller. The IFM is classic direct
inverse modelling \citep{jordan_forward_1992} with hindsight relabelling
\citep{andrychowicz_hindsight_2017}. To our knowledge, no prior work
moves a gait-free footstep planner between two model-based controllers
by refitting only an IFM.

\section{Problem Formulation}
\label{sec:setup}

\noindent\textbf{Setting.}
\looseness=-1 A Unitree Go1 walks forward at a given speed over level
ground with voids. A planner, queried every tick of
$\Delta t = \corlControlTickMs$\,ms, decides the steps, and a
model-based controller $c$ executes them. The planner is given no
gait schedule %
as an input or constraint.

\noindent\textbf{Decisions.}
\looseness=-1 At each tick, the planner takes the no-step option
$\varnothing$ or starts one step \mbox{$(\ell, \mathbf u^\star, T)$},
which lifts leg $\ell$ and should land it at the \emph{desired foothold}
$\mathbf u^\star$ after a swing time $T \in [T_{\min}, T_{\max}] =
[\corlSwingDurationMin, \corlSwingDurationMax]$\,s. %
No more than 2 legs may be in swing at once,
and the desired foothold must be one of the leg's \emph{candidates}
(Sec.~\ref{subsec:method-step}). The planner sees the robot's state, a
\corlHistSeconds\,s history, and the candidates, but neither the %
terrain nor the controller's internal state, so the decision process
is partially observed \citep{yang_fast_2021}.

\noindent\textbf{Closed-loop footstep map.}
\looseness=-1 Given a \emph{command} $\mathbf u$, the foothold sent for
leg $\ell$, and a swing time $T$, the controller lands the foot at
\begin{equation}
  \mathbf o = F_c(\mathbf s, \ell, T, \mathbf u)
    = \mathbf u + \boldsymbol\delta_c(\mathbf s, \ell, T, \mathbf u),
  \label{eq:forward}
\end{equation}
where $\mathbf s$ is everything else that affects the swing. Footholds,
commands, landings, and voids are planar offsets from the stepping
leg's hip at the decision tick, in the gravity-aligned \emph{yaw frame}
along the robot's heading at that tick. Since this frame stays fixed in
the world after the decision tick, the \emph{landing offset}
$\boldsymbol\delta_c$ also contains whatever part of the base's motion
until touchdown the controller passes on to the foot. It is
a property of the controller and differs between controllers, whereas
the \emph{landing error} $\mathbf o - \mathbf u^\star$ compares the
landing with the desired foothold. The execution layer therefore sends
$\mathbf u = I^c(\zeta, \mathbf u^\star)$, where the IFM $I^c$ is meant
to invert $F_c$ and $\zeta$ holds the newest tick of the history, the
leg, and the swing time. The planner has no controller-specific
parameter, but its history reflects the controller, so its inputs can
still shift between controllers.

\noindent\textbf{Problems.}
\looseness=-1 \emph{Problem~1:} with the \emph{training controller}, the
convex MPC of \citet{dicarlo_dynamic_2018} randomized at every reset,
and its IFM fitted beforehand and frozen, find the planner that maximizes the expected sum of rewards over an episode, each discounted in time. \emph{Problem~2:} on the \emph{new controller},
OCS2's NMPC and WBC, keep the planner frozen and fit only its IFM, on
fewer than 20{,}000 OCS2 steps, so that the planner succeeds on held-out
worlds at least as often as OCS2's own gaits. Moving to OCS2 also
changes the simulator, odometry, and terrain source (a depth camera),
and no planner setting is tuned for it (App.~\ref{app:deployment}).

\section{Method}
\label{sec:method}

CADENCE (Fig.~\ref{fig:overview}) turns each candidate footstep, and a
no-step option, into a token from that option and the robot's current
state and recent history; a candidate mixer compares the tokens with each
other, and each gets a score and a swing time
(Sec.~\ref{subsec:method-network}). One option is taken per tick, with
up to two legs in swing. The chosen footstep is sent through the
controller's IFM, and swapping the IFM moves CADENCE to a second
controller without retraining the underlying footstep planning
(Secs.~\ref{subsec:method-idm} and~\ref{subsec:method-training}).

\begin{figure}[t]
  \centering
  \includegraphics[width=0.98\textwidth]{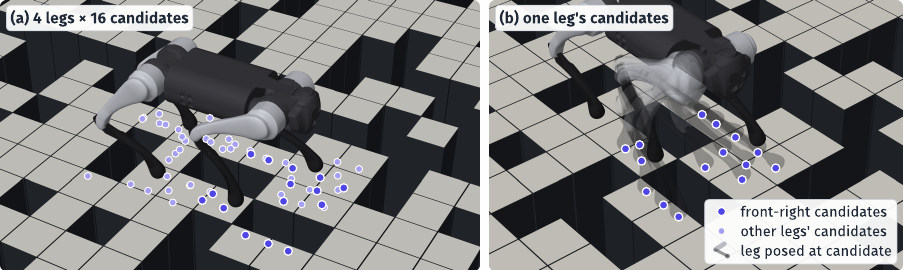}
  \vspace{-1em}
  \caption{The action space at one tick (posed MuJoCo scene,
    illustrative). Each tick, CADENCE takes no step or picks one leg and
    one of its candidates, plus a swing time. (a)~The
    \corlTrainCandidatesPerLeg{} candidates of each leg, by the rule of
    Sec.~\ref{subsec:method-step}; dots the robot hides are not drawn.
    (b)~The front-right leg at each of its candidates.}
\vspace{-1.4em}
  \label{fig:action}
\end{figure}

\subsection{A Small Action Space}
\label{subsec:method-step}

\looseness=-1 At each tick, CADENCE takes one of at most
\corlTotalCandidates{} discrete choices, the no-step option, or one of
\corlTrainCandidatesPerLeg{} candidate footholds for each of the
\corlNumLegs{} legs (Fig.~\ref{fig:action}).
The gait, timing, and order are emergent properties of CADENCE's action
space and sampling rate. The foothold choices contain desired
swing durations, but not joint targets, or gait parameters. The
model-based controller has full control over the swing within the position
and timing constraints set by CADENCE.
The candidate footsteps are chosen from nearby steppable terrain, which we
sample uniformly from safe locations in a 1cm grid with 2cm of clearance added around voids.

\subsection{Network}
\label{subsec:method-network}

\looseness=-1 CADENCE embeds the robot state $\mathbf x_g$, a
\corlHistSeconds\,s history $\mathbf h$ of base velocities, foot
positions, scheduled contacts, and each leg's last \emph{touchdown
error} (its landing minus the command sent), and each candidate's code
$\mathbf q_i$ (its leg and offset from the hip). The state and history
embeddings are added into one context, which is joined with each
candidate's embedding into one token per choice
(Fig.~\ref{fig:overview}a), with a learned embedding for the no-step
option. The \emph{candidate mixer}, a small transformer encoder
\citep{vaswani_attention_2017}, lets the tokens attend to each other, so
every choice, waiting included, is judged against the alternatives. A
shared network then gives each mixed token a score $\alpha_i$ and a
swing time $T_i$, which a logistic function keeps between $T_{\min}$ and
$T_{\max}$, and the policy is a softmax of the scores, sampled in
training and maximized at deployment. With the history's projection and
the mixer's output layers initialized at zero, CADENCE starts out
ranking each candidate in isolation and only introduces mixing as it
learns to (App.~\ref{app:settings}).

\subsection{Inverse Footstep Model}
\label{subsec:method-idm}

\looseness=-1 Our training controller keeps each swing's target fixed
relative to the moving hip, so the foot lands past the command, by up to
\corlOvershootFastMax\,cm at \corlSimSpeedFast\,m/s, about twice the
candidate margin, whereas OCS2 fixes the target in the world. Their footstep
maps $F_c$ (Eq.~\eqref{eq:forward}) therefore differ, and for each
controller the IFM sends a command $\mathbf u = I^c(\zeta,\mathbf
u^\star)$ meant to satisfy $F_c(\mathbf s,\ell,T,\mathbf u) \approx
\mathbf u^\star$. Every logged step $(\zeta,\mathbf u,\mathbf o)$ shows,
in hindsight, a command that landed the foot at $\mathbf o$, so we fit
the inverse directly (Fig.~\ref{fig:overview}b), with the landing as input
\citep{jordan_forward_1992, andrychowicz_hindsight_2017,
akizhanov_learning_2024}:
\begin{equation}
  \min_\theta \; \mathbb E\,
  \big\| \mathbf u - I^c(\zeta, \mathbf o) \big\|^2,
  \qquad
  I^c(\zeta, \mathbf o) = \mathbf o + \Delta_\theta(\zeta, \mathbf o),
  \label{eq:idm-loss}
\end{equation}
where $\Delta_\theta$ is a small multilayer perceptron whose output
layer is initialized at zero, so an unfitted IFM is the identity. For
OCS2, each training seed's frozen CADENCE logs
fewer than 20{,}000 steps through the training controller's IFM, on
worlds separate from the evaluation, and its own OCS2 IFM is fitted on
them.
\phantomsection\label{subsec:idm-theory}Since the IFM does not see the
full state, it can cancel only the part of the landing error that its
inputs predict, and is guaranteed to cancel it only under conditions
that neither deployed IFM meets (App.~\ref{app:idm-proofs}); the
experiments therefore carry the transfer claim
(Sec.~\ref{subsec:exp-why}).

\subsection{Training and Deployment}
\label{subsec:method-training}

\begin{figure}[t]
  \centering
  \includegraphics[width=0.92\textwidth]{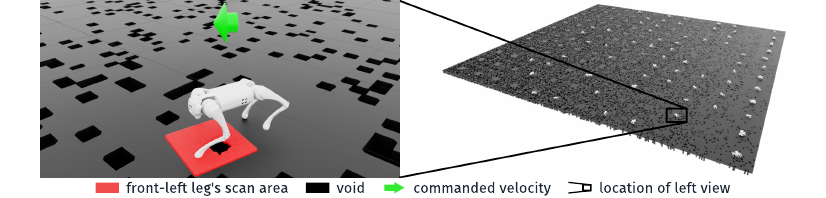}
  \vspace{-0.5em}
  \caption{(Left) One Unitree Go1 on the training terrain. (Right) The
    full training terrain, a grid of $12\times12$ sub-terrains of
    $4\times4$\,m with voids and increasing difficulty.}
  \label{fig:training-terrain-combined}
\end{figure}

\looseness=-1 We train CADENCE with PPO \citep{Schulman_proximal_2017}
in Isaac Lab \citep{mittal_orbit_2023}, on terrain whose void fraction
a curriculum raises to \corlMaxVoidPct\%
(Fig.~\ref{fig:training-terrain-combined}), with the reward
\begin{equation}
  r = \Delta t\,\Big[\,w_a\,\mathbf 1_{\text{alive}}
  + e^{-\|\mathbf v^{\mathrm{usr}}_{xy} - \mathbf v_{b,xy}\|^2/\sigma_v^2}
  + e^{-(\omega^{\mathrm{usr}}_z - \omega_{b,z})^2/\sigma_\omega^2}
  - w_s\,\mathbf 1_{\text{step}}
  - w_f\,\mathbf 1_{\text{fail}}\Big],
  \label{eq:reward}
\end{equation}
which tracks the commanded (usr) base ($b$) velocities and charges
each step and failure but has no gait, clearance, or posture term
(weights $w_a, w_s, w_f$ and widths $\sigma_v, \sigma_\omega$ in
App.~\ref{app:settings}). The training controller's IFM is frozen in the loop (Fig.~\ref{fig:overview}c), and the controller is
randomized at every reset (App.~\ref{app:training}).

\phantomsection\label{subsec:method-deployment}\looseness=-1 On OCS2
(Fig.~\ref{fig:overview}d), CADENCE replaces the gait scheduler of the \texttt{legged\_control}
stack \citep{leggedcontrol, grandia_perceptive_2022} in Gazebo, with
all OCS2 settings stock. Its candidates are built as in training, the
corrected command enters the NMPC as a foot-placement constraint, and
each step joins OCS2's contact schedule at the tick it is chosen.
Terrain comes from a simulated depth camera and odometry from OCS2's
leg-based estimator (App.~\ref{app:deployment}).

\section{Simulation Experiments}
\label{sec:experiments}

\looseness=-1 We test whether CADENCE beats OCS2's own cyclic gaits on
OCS2, what gait emerged and which freedoms of its action space
mattered, and what moving it from the convex MPC to OCS2 took
(Sec.~\ref{subsec:exp-transfer}).

\subsection{Setup}
\label{subsec:exp-setup}

\noindent\textbf{Benchmark.}
\looseness=-1 The benchmark runs on OCS2 in Gazebo
(Sec.~\ref{subsec:method-deployment}) on strips of
\corlBenchTileCm\,cm tiles, each a solid pillar or, with probability
equal to the void fraction, a void. Its six \emph{cells} combine void
fractions of \corlBenchVoidLow, \corlBenchVoidMid, and
\corlBenchVoidHigh\% (the void areas of Isaac's nominal
\corlBenchIsaacVoidLow, \corlBenchIsaacVoidMid, and
\corlBenchIsaacVoidHigh\%) with speeds of \corlBenchSpeedSlow{} and
\corlBenchSpeedFast\,m/s, and each cell holds
\corlBenchTrialsPerCell{} trials of \corlBenchTrialSeconds\,s on the
same worlds for every method. A trial succeeds if the robot does not
fall and walks at least half the nominal distance, the speed times the
trial time. Every learned method is trained with
\corlNumTrainSeeds{} training seeds and evaluated at iteration
\corlOcsCheckpoint, where CADENCE's mean success on the training
controller in Isaac was highest; this iteration was fixed before any
result trials and may not be the best one for the
other methods. We report
the mean, lowest, and
highest seed scores. A sweep adds more void fractions and speeds on other
worlds, with a stricter criterion (App.~\ref{app:sweep}).

\begin{table}[b]
\centering\footnotesize
\caption{Success (\%) on the OCS2 benchmark: mean of
  \corlSeedsGaitNet{} training seeds at iteration \corlOcsCheckpoint{}
  (\corlBenchTrialsPerArm{} trials each), with the lowest and highest
  seed; the stock gaits ran once.}
\label{tab:results}
\begin{tabular}{@{}lrcr@{}}
\toprule
\thd{Method} & \thd{All cells} & \thd{Lowest--highest seed}
  & \thd{\corlBenchVoidHigh\% voids} \\
\midrule
CADENCE & \corlBenchFull & \corlBenchFullMin--\corlBenchFullMax
  & \corlBenchFullDense \\
\midrule
OCS2-Trot & \corlBenchTrot & --- & \corlBenchTrotDense \\
OCS2-Single & \corlBenchStaticWalk & --- & \corlBenchStaticWalkDense \\
\midrule
CADENCE-without-mixer-or-history & \corlBenchMlpMean
  & \corlBenchMlpSeedMin--\corlBenchMlpSeedMax & \corlBenchMlpDense \\
\midrule
CADENCE-Diagonal & \corlBenchDiag & \corlBenchDiagMin--\corlBenchDiagMax
  & \corlBenchDiagDense \\
CADENCE-Single & \corlBenchSingle & \corlBenchSingleMin--\corlBenchSingleMax
  & \corlBenchSingleDense \\
\midrule
CADENCE-with-offset & \corlBenchOffset & \corlBenchOffsetMin--\corlBenchOffsetMax
  & \corlBenchOffsetDense \\
CADENCE-without-IFM & \corlBenchNoIdm
  & \corlBenchNoIdmMin--\corlBenchNoIdmMax & \corlBenchNoIdmDense \\
CADENCE-with-training-IFM & \corlBenchTrainIfm
  & \corlBenchTrainIfmMin--\corlBenchTrainIfmMax & \corlBenchTrainIfmDense \\
\bottomrule
\end{tabular}
\end{table}

\noindent\textbf{Methods.}
\looseness=-1 Here we detail nine methods shown in Table~\ref{tab:results}: CADENCE,
two cyclic gaits from OCS2, and six ablations of CADENCE.
All methods share the depth map, NMPC, and WBC.
OCS2-Trot and OCS2-Single are the stock cyclic gaits
of \citet{leggedperceptive}. The default OCS2 gait parameters were used, having a \corlTrotSwing\,s swing time, a \corlTrotPeriod\,s period for the trot, and
a \corlWalkPeriod\,s period for the static gait (OCS2-Single). OCS2 chooses footholds from planar regions
using a heuristic \citep{grandia_perceptive_2022}.
CADENCE runs with its refitted OCS2 IFM and lets any two legs
swing at once. CADENCE-without-mixer-or-history scores each candidate
alone from the state and its code (Sec.~\ref{subsec:method-network})
and is trained with CADENCE's recipe but with no IFM in the training
loop. CADENCE-Diagonal keeps CADENCE's weights but, at
test time, starts a second swing only on the swinging leg's diagonal
partner, and CADENCE-Single is retrained to swing one leg at a
time. Both CADENCE-without-mixer-or-history and CADENCE-Single run with the OCS2 IFM
of the CADENCE policy with the same training seed, which was fitted on
that policy's steps and not on their own.
Three ablations keep CADENCE's weights and change only the footstep
correction. CADENCE-with-offset adds a constant per-leg offset, the
mean of command minus landing over the OCS2 IFM's logged steps.
CADENCE-without-IFM sends the desired foothold unchanged,
and CADENCE-with-training-IFM uses the training controller's
IFM.

\noindent\textbf{Gait data.}
\looseness=-1 The gait statistics come from the successful trials of an
earlier evaluation run of one CADENCE policy
(\corlGaitTrialsGaitNet{} of \corlBenchTrialsPerArm) and of the trot
(\corlGaitTrialsTrot) on other worlds, not from the runs of
Table~\ref{tab:results} (App.~\ref{app:earlier-run}). A fixed period
gives a near-zero coefficient of variation (CV) of a leg's time between
lift-offs, and a fixed order that lifts each leg once per cycle, at
most two at a time, never repeats a \emph{step event} (the legs lifting
within \corlGaitEventWindowMs\,ms of its first lift-off) immediately,
nor lifts a leg twice in a row.

\noindent\textbf{Metrics.}
\looseness=-1 We report success rates and their differences in
percentage points, and, on flat ground at \corlBenchSpeedSlow{} and
\corlBenchSpeedFast\,m/s, OCS2's RMS landing error, the distance from
the desired foothold to the landing.

\begin{figure*}[!b]
  \centering
  \includegraphics[width=\textwidth]{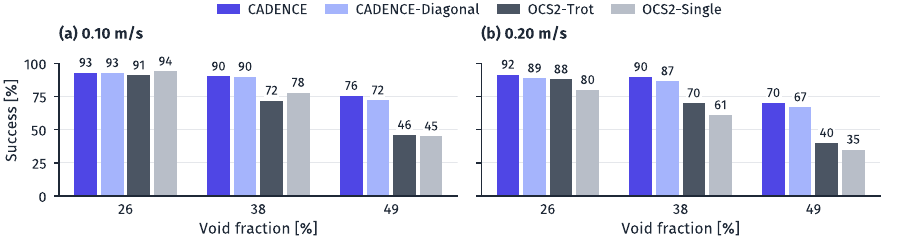}
  \vspace{-2em}
  \caption{Success on the OCS2 benchmark per cell, at
    (a)~\corlBenchSpeedSlow\,m/s and (b)~\corlBenchSpeedFast\,m/s,
    \corlBenchTrialsPerCell{} trials per bar and training seed. Learned
    methods: mean of \corlNumTrainSeeds{} training seeds, in whole
    percent. Overall success is in Table~\ref{tab:results}.}
  \label{fig:bench}
\end{figure*}

\subsection{Results}
\label{subsec:exp-transfer}

\noindent\textbf{CADENCE beat OCS2's cyclic gaits, most where
voids are densest (Fig.~\ref{fig:bench}, Table~\ref{tab:results}).}
\looseness=-1 CADENCE succeeded in \corlBenchFull\% of benchmark
trials, against \corlBenchTrot\% for OCS2-Trot and
\corlBenchStaticWalk\% for OCS2-Single, leads of 17.3 and 19.6
percentage points, and even its lowest seed led the trot by 10.5. The
lead over the trot grew with the void fraction, from 2 to 18 to 30
points at \corlBenchSpeedSlow\,m/s and from 4 to 20 to 30 at
\corlBenchSpeedFast\,m/s, although the walk beat CADENCE in the easiest
cell (94 against 93\%). In the hardest cell, \corlBenchVoidHigh\% voids
at \corlBenchSpeedFast\,m/s, CADENCE succeeded in 70\% against 40\% for
the trot and 35\% for the walk (Fig.~\ref{fig:bench}b). CADENCE-without-mixer-or-history succeeded in
\corlBenchMlpMean\%, 13.8 points below CADENCE, from which it differs in
the mixer, the history, network size, the IFM in training, and an OCS2
IFM fitted on CADENCE's steps. Since the stock gaits
also take footholds from another source, the lead belongs to CADENCE and
its execution layer together.

\begin{figure*}[t]
  \centering
  \includegraphics[width=0.94\textwidth]{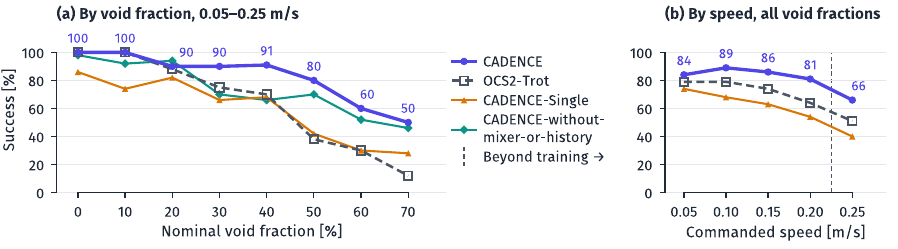}
  \vspace{-1em}
  \caption{Sweep on OCS2, on worlds other than the benchmark's, with a
    stricter criterion (reach the end of a \corlOcsTrackM\,m track in
    time); \corlOcsTrialsPerPoint{} trials per point and training
    seed; learned methods: mean of \corlNumTrainSeeds{} seeds. (a)~By nominal void
    fraction, speeds mixed; above \corlMaxVoidPct\%, beyond the training
    curriculum. (b)~By speed; right of the dashed line, above the
    training speeds.}
  \label{fig:sweep-main}\label{fig:sweep}
\end{figure*}

\looseness=-1 The sweep, on other worlds and with a criterion that also
fails a robot that stalls, extends this result beyond training. In
Fig.~\ref{fig:sweep-main}a, CADENCE reached \corlSweepFull\% against
the trot's \corlSweepTrot\%. The two were
equal up to a nominal 10\%, and from 30\% on CADENCE led by 15 to 42
points. Beyond the \corlMaxVoidPct\% curriculum, at 50, 60, and
\corlOcsMaxVoidPct\% voids, it succeeded in 80, 60, and 50\% against
the trot's 38, 30, and 12\%. It also led at every speed, and at
\corlOcsSpeedMax\,m/s, above the training speeds, it succeeded in 66
against 51\%. CADENCE therefore kept its lead on unseen worlds, denser
voids, and a higher speed than it was trained on.

\begin{figure*}[t]
  \centering
  \includegraphics[width=0.94\textwidth]{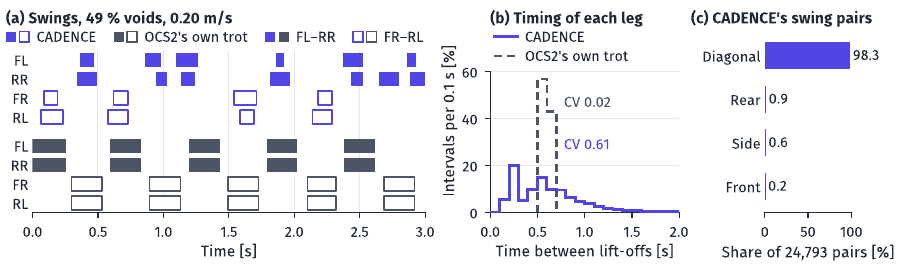}
  \vspace{-1em}
  \caption{Gait of CADENCE and OCS2-Trot (earlier evaluation run,
    successful trials). (a)~Swings in one trial each (world seed
    \corlGaitExampleSeed, \corlBenchVoidHigh\% voids,
    \corlBenchSpeedFast\,m/s; FL:~front left, RR:~rear right).
    (b)~Time between a leg's lift-offs. (c)~Leg pairs of overlapping
    swings.}
\vspace{-1.4em}
  \label{fig:gait}
\end{figure*}

\begin{figure}[t]
  \centering
  \subfigure[Measured swings][t]{%
    \includegraphics[width=0.46\textwidth]{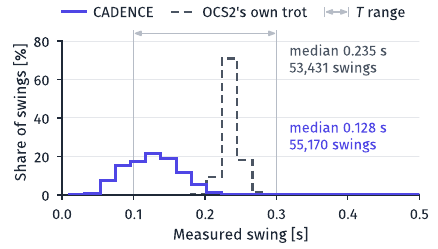}}\hfill
  \subfigure[Footstep corrections on OCS2][t]{%
    \includegraphics[width=0.46\textwidth]{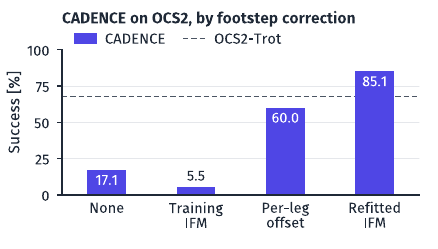}}
  \caption{(a)~Measured swings, earlier run ($T$ range: commanded swing
    times); the detector shortens every swing, so only the comparison
    holds. (b)~CADENCE on OCS2 with four footstep corrections, mean of
    \corlSeedsGaitNet{} seeds.}
\vspace{-1.4em}
  \label{fig:swing-transfer}
\end{figure}

\noindent\textbf{The emergent gait is acyclic and often pairs diagonal legs
(Figs.~\ref{fig:gait} and~\ref{fig:swing-transfer}a).}
\looseness=-1 With no gait schedule, gait reward, or phase variable,
CADENCE's gait had neither a fixed period nor a fixed order. In the
example trial of Fig.~\ref{fig:gait}a, the trot alternated its two
diagonal pairs on its fixed period, whereas CADENCE spaced its steps
unevenly and sometimes split a pair into single-leg steps. Over all
successful trials, a leg's time between lift-offs varied far more than
in the trot (CV \corlGaitStrideCv{} against \corlGaitStrideCvTrot;
Fig.~\ref{fig:gait}b), and CADENCE raised its step rate with speed, from
\corlGaitStepsSlow{} steps per second at \corlBenchSpeedSlow\,m/s to
\corlGaitStepsFast{} at \corlBenchSpeedFast\,m/s, while the trot kept
\corlGaitStepsTrot{} at both. A step event repeated the legs of the
one before it in \corlGaitEventRepeatPct\% of consecutive events, and a
leg lifted twice in a row in \corlGaitLegTwicePct\% of consecutive
lift-offs, against \corlGaitEventRepeatPctTrot{} and
\corlGaitLegTwicePctTrot\% for the trot, and four consecutive lift-offs
covered all four legs in \corlGaitFourAllPct{} against
\corlGaitFourAllPctTrot\%. CADENCE's measured swings were shorter
(median \corlGaitSwingMedian{} against \corlGaitSwingMedianTrot\,s;
Fig.~\ref{fig:swing-transfer}a). Yet \corlGaitDiagPairPct\% of its
\corlGaitOverlapPairs{} overlapping swing pairs were diagonal
(Fig.~\ref{fig:gait}c), with partners lifting off a median
\corlGaitLiftLagMs\,ms apart, about one tick. CADENCE therefore kept
the trot's pairing but neither its period nor its order, in this one
run of one policy.

\looseness=-1 Restricting the action space tested which freedom
mattered. Retrained to swing one leg at a time, CADENCE-Single
succeeded in only \corlBenchSingle\%, though OCS2-Single, also
one leg at a time, reached \corlBenchStaticWalk\%. This loss mixes the
missing overlap with an OCS2 IFM fitted on CADENCE's two-leg steps. Restricted at test time to diagonal
pairs, CADENCE-Diagonal lost only 2.2 points on the benchmark (1.2 to
2.8 in each seed), but 9.4 in the sweep (\corlSweepDiag\%, not shown).

\noindent\phantomsection\label{subsec:exp-why}\textbf{Moving CADENCE to
OCS2 took a refitted IFM model (Fig.~\ref{fig:swing-transfer}b).}
\looseness=-1 With CADENCE's weights frozen, it succeeded in
\corlBenchNoIdm\% without the IFM and in only \corlBenchTrainIfm\% with
the training controller's IFM, consistent with that IFM cancelling a
hip-fixed overshoot that OCS2's world-fixed target lacks
(Sec.~\ref{subsec:method-idm}). A constant per-leg offset recovered much
of the loss (\corlBenchOffset\%), and the refitted IFM added 25.1
points more; we did not test models between the two. Overall, only the
refit lifted CADENCE above the trot, although at \corlBenchVoidHigh\%
voids the offset also beat it (54 and 47 against 46 and 40\%).
On flat ground, the refit cut OCS2's RMS landing error from
about \corlOcsLandRmsNoIdm\,cm, three times the \corlMarginCm\,cm
margin, to about \corlOcsLandRmsIdm\,cm (Fig.~\ref{fig:teaser}c).

\section{Conclusion and Future Work}
\label{sec:conclusion}
\looseness=-1 We presented CADENCE, a learned step planner with a small
action space for model-based controllers. Every
\corlControlTickMs\,ms, it picks one of at most \corlTotalCandidates{}
choices, no step, or one leg's step to a safe foothold, and
predicts a swing time. Nothing in training prescribed a gait, yet the
policy we analyzed developed an acyclic gait, with no fixed period or
stepping order, that still paired diagonal legs. In trials across varied terrain (26\% to 49\% voids) and speeds (0.1\,m/s and 0.2\,m/s), and with only its IFM
refitted to run on OCS2's NMPC, CADENCE succeeded in
\corlBenchFull\% of trials, against \corlBenchTrot\% for
OCS2-Trot and \corlBenchStaticWalk\% for OCS2-Single. At \corlBenchVoidHigh\% voids,
CADENCE succeeded in \corlBenchFullDense\% of trials against OCS2-Trot's \corlBenchTrotDense\%. Retrained to swing one leg at a time, it
reached only \corlBenchSingle\%, and no simpler footstep correction lifted CADENCE
above OCS2-Trot. The baselines were OCS2's untuned gaits, acyclicity was measured on one earlier run of one policy,
and the refit needed steps logged on the new controller.

\looseness=-1 Future work will move CADENCE to additional
model-based controllers, measure how few logged steps the refit needs,
and test whether relogging with the deployed IFM
\citep{christiano_transfer_2016} or online adaptation
\citep{xie_fada_2026} reduces the remaining landing error. We will also
extend the candidates to stairs and slopes and compare with gait-free planners.

\vspace{1.5em}
\par\smallskip\noindent\textbf{Generative AI Use.}
Generative AI tools were used to search the literature, check
arithmetic and derivations, assist with coding the
experiments and plotting scripts, and draft and edit the text. The authors
reviewed and verified all AI-assisted content and take responsibility for the paper.

\bibliography{refs}

\appendix
\clearpage
\section{Settings}
\label{app:settings}

\noindent\textbf{Action space and network.}
\looseness=-1 The candidate grid spans
$\pm\corlGridReachCm$\,cm around the hip in \corlScanResCm\,cm cells,
and the unsafe region is grown by two cells, so no unsafe ray lies
within the candidate margin $m = \corlMarginCm$\,cm of a remaining safe
cell along the grid axes. The history covers $H = \corlHistTicks$
ticks. The candidate mixer has \corlMixerLayers{} layers of width
\corlMixerDim, \corlMixerHeads{} heads, and feed-forward blocks of
width \corlMixerFF. CADENCE has
\corlParamsGaitNetK{}k parameters, and CADENCE-without-mixer-or-history
(Sec.~\ref{subsec:exp-setup}) has \corlParamsPerCandidateK{}k.
Base and commanded velocities and the history's hip-relative foot
positions are in the base frame, and candidates are in each hip's yaw
frame (Sec.~\ref{sec:setup}).

\noindent\textbf{IFM of the training controller.}
\looseness=-1 Its network $\Delta_\theta$ (Eq.~\eqref{eq:idm-loss}) is
a $2 \times \corlIdmHidden$ MLP fitted on \corlIdmFootsteps{} steps,
logged at up to 30\% voids and over our speed range, with Gaussian
noise on each command to widen the data. On a held-out \corlIdmHeldOutPct\% of the steps, its error
is \corlIdmRmsCm\,cm RMS, against \corlIdmIdentityRmsCm\,cm for the
identity map. Without correction, the training controller lands the
foot +\corlOvershootSlowMin{} to +\corlOvershootSlowMax\,cm past the
command along the walking direction at \corlSimSpeedSlow\,m/s and
+\corlOvershootFastMin{} to +\corlOvershootFastMax\,cm at
\corlSimSpeedFast\,m/s.

\noindent\textbf{Training (Eq.~\eqref{eq:reward}).}
\looseness=-1 PPO runs \corlParallelEnvs{} robots with rollouts of
\corlRolloutSteps{} ticks, a learning rate of $\corlLearningRate$,
\corlPpoEpochs{} epochs, \corlMiniBatches{} minibatches, a clip of
\corlClip, an entropy coefficient of \corlEntropyCoef, a discount factor $\gamma = \corlGamma$, and a generalized-advantage parameter $\lambda = \corlLambda$. The reward has weights $w_a =
\corlAliveWeight$, $w_s = \corlActionPenalty$, and $w_f =
\corlTerminationPenalty$ and tracking widths $\sigma_v =
\corlTrackXyStd$\,m/s and $\sigma_\omega = \corlTrackYawStd$\,rad/s,
and swing times are drawn with standard deviation $\sigma_T =
\corlDurationExploreStd$\,s. Commanded forward speeds span
0--\corlCmdVxMax\,m/s and yaw rates $\pm\corlCmdYawMax$\,rad/s. Tiles
are \corlTerrainCellCm\,cm with voids \corlVoidDepth\,m deep, and the
\corlNumLevels{} curriculum levels span 0--\corlMaxVoidPct\% voids. The
training controller's MPC runs at \corlMpcHz\,Hz over a
\corlMpcHorizon\,s horizon, with leg torques at \corlPhysicsHz\,Hz.

\noindent\textbf{OCS2 IFM and footstep corrections.}
\looseness=-1 Each training seed has its own OCS2 IFM, with the network
and fit of the training controller's IFM, fitted on fewer than 20{,}000 steps that this seed's frozen CADENCE
took on OCS2 through the training controller's IFM, on worlds disjoint
from the evaluation's. Both CADENCE-without-mixer-or-history and CADENCE-Single run
with the OCS2 IFM of CADENCE's same training seed. Table~\ref{tab:percell} gives the success of every
method, the controls included, in each benchmark cell.

\noindent\textbf{Sweep.}\phantomsection\label{app:sweep}
\looseness=-1 The sweep (Fig.~\ref{fig:sweep-main}) covers eight void
fractions from 0 to \corlOcsMaxVoidPct\% and five speeds
from \corlOcsSpeedMin{} to \corlOcsSpeedMax\,m/s, on Isaac's
nominal scale, on which the benchmark's \corlBenchVoidLow,
\corlBenchVoidMid, and \corlBenchVoidHigh\% are \corlBenchIsaacVoidLow,
\corlBenchIsaacVoidMid, and \corlBenchIsaacVoidHigh\%, with
the benchmark's checkpoints, on worlds other than the
benchmark's, shared by all methods. A trial succeeds only if the robot
reaches the end of a \corlOcsTrackM\,m track within \corlOcsTimeFactor{}
times the time needed at the commanded speed. The two panels are separate run sets with
\corlOcsTrialsPerPoint{} trials per point, spread over the five speeds
at each void fraction in panel~(a) and 10 per void fraction at each
speed in panel~(b).

\begin{table}[t]
\centering\footnotesize
\caption{Success (\%) per cell of the OCS2 benchmark, in whole percent
  (Table~\ref{tab:results} pools the unrounded values). Learned methods:
  mean of \corlSeedsGaitNet{} training seeds at iteration
  \corlOcsCheckpoint, \corlBenchTrialsPerCell{} trials per cell and
  seed. The stock gaits ran once.}
\label{tab:percell}
\begin{tabular}{@{}lrrrrrr@{}}
\toprule
 & \multicolumn{3}{c}{\thd{\corlBenchSpeedSlow\,m/s}}
 & \multicolumn{3}{c}{\thd{\corlBenchSpeedFast\,m/s}} \\
\cmidrule(lr){2-4}\cmidrule(l){5-7}
\thd{Method} & \thd{\corlBenchVoidLow\%} & \thd{\corlBenchVoidMid\%}
  & \thd{\corlBenchVoidHigh\%} & \thd{\corlBenchVoidLow\%}
  & \thd{\corlBenchVoidMid\%} & \thd{\corlBenchVoidHigh\%} \\
\midrule
CADENCE                 & 93 & 90 & 76 & 92 & 90 & 70 \\
\midrule
OCS2-Trot               & 91 & 72 & 46 & 88 & 70 & 40 \\
OCS2-Single             & 94 & 78 & 45 & 80 & 61 & 35 \\
\midrule
CADENCE-without-mixer-or-history & 81 & 79 & 69 & 73 & 71 & 54 \\
\midrule
CADENCE-Diagonal        & 93 & 90 & 72 & 89 & 87 & 67 \\
CADENCE-Single          & 45 & 45 & 40 & 38 & 38 & 31 \\
\midrule
CADENCE-with-offset        & 67 & 66 & 54 & 64 & 63 & 47 \\
CADENCE-without-IFM & 21 & 20 & 14 & 18 & 17 & 12 \\
CADENCE-with-training-IFM  &  8 &  6 &  7 &  6 &  4 &  2 \\
\bottomrule
\end{tabular}
\end{table}

\section{Training Details}
\label{app:training}

\noindent\textbf{Policy and critic.}
\looseness=-1 The action is a choice $i$ among \corlTotalCandidates{}
\emph{slots}, each a leg and candidate, the no-step option, or empty,
and a swing time $T$. With $\mathbf x$ the planner's input, the policy
factors as $\pi(i, T \mid \mathbf x) = \pi(i \mid \mathbf x)\,\mathcal
N(T;\, T_i(\mathbf x), \sigma_T^2)$, with $\pi(i \mid \mathbf x)$ the
softmax of the scores $\alpha_i$ over the non-empty slots, $T_i$ the
swing time of Sec.~\ref{subsec:method-network}, and $\sigma_T$ fixed
(App.~\ref{app:settings}). The sampled swing time is not clipped, so in
training it can fall outside $[T_{\min}, T_{\max}]$, whereas deployment
uses the mean $T_i$. The critic has no attention: an MLP on the state embedding and each slot's candidate
embedding gives one value per slot, with one learned embedding for the
no-step option and the empty slots, and a second MLP
($\corlTotalCandidates \to 32 \to 32 \to 1$) maps the slot values to
the value estimate. It reads the slots in a fixed order, each leg's
first slot holding the pinned nominal foothold
(App.~\ref{app:sampler}).

\noindent\textbf{Episodes and failure.}
\looseness=-1 An episode lasts up to \corlEpisodeSeconds\,s and fails
earlier if the trunk tilts by more than \corlTiltLimitDeg$^\circ$, drops
below \corlMinBaseHeight\,m, or a foot drops more than
\corlFootDepthLimit\,m below the ground. Commanded velocities are
forward-only, since the robot maps terrain with a forward-facing
camera, and the tracking widths are narrow, since with wider ones a
standing robot would earn most of the tracking reward at our speeds.
Observations are not normalized and carry no added noise, and there are
no external pushes. Each training seed took about
\corlTrainHours\,h to reach iteration \corlOcsCheckpoint{} on
one NVIDIA RTX PRO 6000 Blackwell GPU.

\noindent\textbf{Curriculum.}
\looseness=-1 Higher levels have more voids, and robots start on the
three lowest levels. A robot progresses if it walks at least
\corlCurriculumProgress{} times the commanded distance, and it moves up
a level with probability $p_{\text{up}}$ after a progressing time-out
and down with probability $p_{\text{down}} = \corlWalkPDown$ after a
failure or a stalled time-out. The share of failed or stalled episodes
therefore settles near $p_{\text{up}}/(p_{\text{up}} +
p_{\text{down}})$, which is $1/4$ for $p_{\text{up}} =
\corlWalkPUp$. A further \corlWalkRandomPct\% of episodes run on a
random level and do not move the robot's level.

\noindent\textbf{Controller randomization.}
\looseness=-1 At every reset, each robot's controller is redrawn, with
a scale of \corlRandTrackMin{} to 1 on the velocity its stance MPC
tracks, a scale of \corlRandPlanarMin{} to 1 on its planar tracking
weights, a swing apex height of \corlRandApexMin{} to
\corlRandApexMax\,m, each drawn uniformly, a constant per-leg landing
bias of up to $\pm\corlRandBiasCm$\,cm per axis, Gaussian landing noise
of standard deviation \corlRandNoiseCm\,cm, and a lift-off delay
of zero or one tick, the only case in which a step starts after the
tick it is chosen. Every chosen candidate is corrected by the IFM
before this controller executes it, and CADENCE's touchdown errors are
measured against the IFM's output, so its history reports how far the
controller missed the corrected command.

\section{Deployment Details}
\label{app:deployment}

\noindent\textbf{Steps and footholds.}
\looseness=-1 On OCS2, CADENCE is queried every tick, also while two legs
swing, and each step it starts joins those already in OCS2's contact
schedule, at absolute times on the NMPC clock. The OCS2 IFM's
command is projected onto the nearest planar region of the segmented
terrain map \citep{grandia_perceptive_2022}, and the NMPC constrains
the foot to land inside that region, in a box of
$\pm$\corlAuthorityCm\,cm around the projected point. OCS2's swing
planner receives only the touchdown heights.

\noindent\textbf{Terrain and odometry.}
\looseness=-1 CADENCE and the controller get terrain from a simulated,
noise-free RealSense D435 depth camera through a depth-vote map, a fine
grid whose squares each depth frame votes hit or miss. A square is
walkable while its fading vote total is positive and it lies far enough
from unwalkable ones, a distance that grows with the distance walked
since the robot last saw it. Odometry is OCS2's leg-based state
estimate with a slip gate, which ignores feet whose velocity disagrees
with it, and the base's pose reference, which integrates the commanded
velocity, stays a short distance ahead of the base.

\section{Analysis of the Inverse Footstep Model}
\label{app:idm-proofs}

\noindent\textbf{Mean error and residual.}
\looseness=-1 The IFM sees $\zeta$ but not the full state $\mathbf s$, so
given $\zeta$ and the command $\mathbf u$ the landing $\mathbf o$ is random,
even though $F_c$ is deterministic. With $\bar F_c(\zeta,\mathbf u) :=
\mathbb E[\mathbf o \mid \zeta,\mathbf u]$, the landing and the landing
error of the command $\mathbf u = I^c(\zeta,\mathbf u^\star)$ split as
\begin{equation}
  \mathbf o = \bar F_c(\zeta,\mathbf u) + \boldsymbol\eta, \qquad
  \mathbf e := \mathbf o - \mathbf u^\star
    = \boldsymbol\beta + \boldsymbol\eta, \qquad
  \boldsymbol\beta := \bar F_c(\zeta,\mathbf u) - \mathbf u^\star ,
  \label{eq:mean-forward}
\end{equation}
where the residual $\boldsymbol\eta$ has mean zero given $(\zeta,\mathbf
u)$, and so given the IFM's inputs $(\zeta,\mathbf u^\star)$ when
$I^c(\zeta,\cdot)$ is one-to-one. Such an IFM therefore acts only on the
mean error $\boldsymbol\beta$, the part of the landing error that its
inputs predict, and none removes $\boldsymbol\eta$. This appendix states
when the regression of Eq.~\eqref{eq:idm-loss} cancels $\boldsymbol\beta$
and which bias remains otherwise.

\noindent\textbf{Notation.}
\looseness=-1 We drop the superscript $c$ from $I^c$. Let $P_c$ be the
distribution of the logged tuples $(\zeta,\mathbf u,\mathbf o)$, one per
step the controller executed, with $\mathbb E_{P_c}(\|\mathbf u\|^2 +
\|\mathbf o\|^2) < \infty$, and let $I_\star(\zeta,\mathbf o) := \mathbb
E_{P_c}[\mathbf u \mid \zeta,\mathbf o]$ be the function that the
regression estimates. Let $\mathcal S$ be the support of the logged pairs
$(\zeta,\mathbf u)$. \emph{Deployment} means any use of a fitted IFM, in
training or on OCS2, and a \emph{deployment pair} is a pair
$(\zeta,\mathbf u^\star)$ at which it is queried. In
Assumption~\ref{as:idm-exact}, $\bar F_c$ is the conditional mean under
$P_c$; in Eq.~\eqref{eq:mean-forward} it is taken under the law of the
deployed steps, and (i) makes the two agree. Norms are Euclidean.

\begin{assumption}[Exact relabelling]
\label{as:idm-exact}
(i)~\emph{Same landing law:} one kernel gives the law of $\mathbf o$ at
every $(\zeta,\mathbf u) \in \mathcal S$, when logging and at deployment.
(ii)~\emph{No residual:} at every $(\zeta,\mathbf u) \in \mathcal S$ this
law is the point mass at $\bar F_c(\zeta,\mathbf u)$.
(iii)~\emph{One command per landing:} for every $\zeta$, the map $\mathbf
u \mapsto \bar F_c(\zeta,\mathbf u)$ is one-to-one on the commands with
$(\zeta,\mathbf u) \in \mathcal S$.
(iv)~\emph{Coverage:} every deployment pair $(\zeta,\mathbf u^\star)$ lies
in the support of the logged pairs $(\zeta,\mathbf o)$.
(v)~\emph{Exact fit:} the fitted IFM $I$ is continuous and equals
$I_\star$ $P_c$-almost everywhere, and $\bar F_c$ (on $\mathcal S$) and
$I_\star$ have continuous versions, which we use.
\end{assumption}

Condition (iii) holds whenever the mean landing offset changes less than
the command, that is, when $\mathbf u \mapsto \bar F_c(\zeta,\mathbf u) -
\mathbf u$ is Lipschitz with a constant below one.

\begin{proposition}[Hindsight regression]
\label{prop:idm-mean}
(a)~Among the square-integrable functions of $(\zeta,\mathbf o)$,
$I_\star$ minimizes the loss of Eq.~\eqref{eq:idm-loss}, uniquely up to
$P_c$-null sets.
(b)~Under Assumption~\ref{as:idm-exact}, every deployment pair has $\bar
F_c(\zeta, I(\zeta,\mathbf u^\star)) = \mathbf u^\star$, so the foot lands
at $\mathbf u^\star$.
\end{proposition}

\begin{idmproof}{Proposition~\ref{prop:idm-mean}}
(a)~Since $I_\star - I$ depends only on $(\zeta,\mathbf o)$ and $\mathbb
E[\mathbf u - I_\star \mid \zeta,\mathbf o] = \mathbf 0$, the cross term
vanishes and $\mathbb E\|\mathbf u - I\|^2 = \mathbb E\|\mathbf u -
I_\star\|^2 + \mathbb E\|I_\star - I\|^2$, which is smallest exactly when
$I = I_\star$ almost everywhere.

(b)~We use twice that a closed set of probability one contains the
support, the smallest such set. By (ii), a logged step lands at $\bar
F_c(\zeta,\mathbf u)$ with $(\zeta,\mathbf u) \in \mathcal S$, and by (iii)
only one such command reaches a given $(\zeta,\mathbf o)$. The logged
command is therefore almost surely a measurable function of
$(\zeta,\mathbf o)$, since a continuous one-to-one map on the closed set
$\mathcal S$ has a Borel inverse, so it equals its conditional mean, and $(\zeta, I_\star(\zeta,\mathbf o)) \in
\mathcal S$ with $\bar F_c(\zeta, I_\star(\zeta,\mathbf o)) = \mathbf o$
for $P_c$-almost every $(\zeta,\mathbf o)$. The set where both hold is
closed, since $\mathcal S$ is closed and $\bar F_c$ and $I_\star$ are
continuous by (v). It therefore contains the support of the logged
$(\zeta,\mathbf o)$ and, by (iv), every deployment pair, so $\bar
F_c(\zeta, I_\star(\zeta,\mathbf u^\star)) = \mathbf u^\star$. By (v), the
closed set where $I$ and $I_\star$ agree also has probability one, hence
contains that support, and $I(\zeta,\mathbf u^\star) =
I_\star(\zeta,\mathbf u^\star)$. By (i) and (ii), the foot lands at
$\bar F_c(\zeta, I(\zeta,\mathbf u^\star)) = \mathbf u^\star$ at
deployment.
\end{idmproof}

\noindent\textbf{Why conditions (ii)--(iv) matter.}
\looseness=-1 Without (ii), the regression averages the command over
hidden states that share $(\zeta,\mathbf o)$, so it is generally biased
even with unlimited data \citep{eysenbach_imitating_2022}. For example,
keep (i) and (v): if, given $\zeta$, $\bar F_c$ is affine in $\mathbf u$
with slope $A$ and the logged commands and the residual are jointly
Gaussian with covariances $\Sigma_u$ and $\Sigma_\eta \succ 0$, the mean
landing is $\bar{\mathbf o} + \Lambda(\mathbf u^\star - \bar{\mathbf o})$,
where $\bar{\mathbf o}$ is the mean logged landing for that $\zeta$ and
$\Lambda = A\Sigma_u A^\top (A\Sigma_u A^\top + \Sigma_\eta)^{-1}$ has
eigenvalues in $[0,1)$: the attenuation bias of a regression on a
noisy input \citep{fuller_measurement_1987}.
Without (iii), two commands with one landing can average to
one that lands elsewhere \citep{jordan_forward_1992}, and without (iv), the held-out error is
silent on targets outside the logged landings.

\noindent\textbf{The deployed IFMs.}
\looseness=-1 Neither deployed IFM meets the assumption. Both leave a
residual, since $\zeta$ omits the terrain and the controller's state, so
(ii) fails, and neither fit is exact, so (v) fails. On OCS2, (iii) can
also fail, since a command outside a planar region is projected onto
it (App.~\ref{app:deployment}). The training
controller's IFM also departs from (i), since it is fitted without the
controller randomization it runs under
(Sec.~\ref{subsec:method-training}), and the OCS2 IFM from (iv), since
its logged landings cover the footholds CADENCE reached through the
training controller's IFM, not the deployed targets. Even on one
controller, (i) can fail, since the logging policy and
CADENCE choose commands from terrain that $\zeta$ omits.
Sec.~\ref{subsec:exp-why} therefore tests the refit against controls
rather than rely on Proposition~\ref{prop:idm-mean}.

\section{Candidate Sampling}
\label{app:sampler}

\looseness=-1 Every tick, for each leg that may step, CADENCE places
its candidates to cover the leg's safe cells, instead of drawing them at
random. Let $S$ be the safe cells left on the leg's grid once the
candidate margin is removed (Sec.~\ref{subsec:method-step}), in the
hip's yaw frame. The sampler returns $K =
\corlTrainCandidatesPerLeg$ distinct safe cells, the candidates $Q =
\{\mathbf q_0, \dots, \mathbf q_{K-1}\}$, or all safe cells if there are
fewer, leaving the other slots empty. It aims at a low quantization
energy $E(Q) = |S|^{-1} \sum_{\mathbf c \in S} \min_i \|\mathbf c -
\mathbf q_i\|^2$, the mean squared distance from a safe cell to its
nearest candidate, which is the $k$-means objective on the safe cells.
Since its minimizer has no closed form, the sampler pins one candidate,
spreads the others, and refines them.

\noindent\textbf{Pinned nominal foothold.}
\looseness=-1 The first candidate is the safe cell nearest the foothold
of a regular stride, $\mathbf q_0 = \arg\min_{\mathbf c \in S} \|\mathbf
c - \mathbf q_{\text{nom}}\|$ with $\mathbf q_{\text{nom}} =
T_{\text{nom}}\, \mathbf v^{\mathrm{usr}}_{xy} + (0,\ y_{\text{foot}} -
y_{\text{hip}})$, whose first term follows the commanded base velocity
$\mathbf v^{\mathrm{usr}}_{xy}$ of Eq.~\eqref{eq:reward} and whose
second keeps the foot's current lateral distance from the hip. The
horizon $T_{\text{nom}} = \corlNomHorizonS$\,s is a fixed prior, not a
fitted value: \corlNomLiftS\,s to lift-off, a \corlNomSwingS\,s swing,
and half of a stance as long as the swing, which puts $\mathbf
q_{\text{nom}}$ at the neutral point of a symmetric stance at
touchdown. Steps still start at the tick they are chosen. The first
candidate does not move afterwards, and, when more than $K$ cells are
safe, it fills the first slot of its leg (App.~\ref{app:training}).

\noindent\textbf{Spread.}
\looseness=-1 Farthest-point sampling
\citep{gonzalez_clustering_1985} adds the others: $\mathbf q_k$ is the
safe cell farthest from $\mathbf q_0, \dots, \mathbf q_{k-1}$, for $k =
1, \dots, K-1$. Most of these picks lie on the boundary of the safe
region.

\noindent\textbf{Refinement.}
\looseness=-1 \Numberstringnum{\corlAnnealSteps} steps of deterministic annealing
\citep{rose_deterministic_1998}, a soft $k$-means on $E$, then pull the
candidates inward. At step $t = 0, \dots, 3$, the temperature is
$\tau_t = \tau_0 (\tau_{\text{end}} / \tau_0)^{t/3}$, with $\tau_0 =
(\corlAnnealTauStartCm\,\text{cm})^2$ and $\tau_{\text{end}} =
(\corlAnnealTauEndCm\,\text{cm})^2$. Each safe cell $\mathbf c_j$ is
shared among the candidates with softmax weights $p_{ij} \propto
\exp(-\|\mathbf c_j - \mathbf q_i\|^2 / \tau_t)$, normalized over $i$,
and each candidate but $\mathbf q_0$ moves to the mean of its share,
$\tilde{\mathbf q}_i = \sum_j p_{ij}\, \mathbf c_j / \sum_j p_{ij}$.
After each step, every target $\tilde{\mathbf q}_i$ returns to the
nearest safe cell. When two candidates claim one cell, the one added
earlier keeps it and the other takes the nearest free cell.

\noindent\textbf{Mirror symmetry.}
\looseness=-1 Ties go to the lowest cell index, and the left legs use
the mirrored problem, so mirrored inputs give mirrored candidates.

\section{Earlier Evaluation Run: Gait Statistics}
\label{app:earlier-run}

\noindent\textbf{The run.}
\looseness=-1 The gait statistics (Figs.~\ref{fig:gait}
and~\ref{fig:swing-transfer}a) come from an earlier evaluation run on
OCS2, not from the runs of Table~\ref{tab:results}. It ran one CADENCE
policy, deployed as in Sec.~\ref{subsec:method-deployment} with an OCS2
IFM and depth-camera terrain, and OCS2-Trot, on other worlds than
Table~\ref{tab:results} but with the benchmark's six cells,
\corlBenchTrialsPerCell{} trials of \corlBenchTrialSeconds\,s per cell,
and success criterion. CADENCE succeeded in \corlGaitTrialsGaitNet{} of
\corlBenchTrialsPerArm{} trials (91.5\%) and the trot in
\corlGaitTrialsTrot{} (84.0\%). Both succeeded more often than in
Table~\ref{tab:results}, the trot by far more (\corlBenchTrot\% there),
so CADENCE's lead was 7.5 points in this run against 17.3 in
Table~\ref{tab:results}. No success rate in the main text comes from
this run.

\noindent\textbf{Swing detection.}
\looseness=-1 The run's log holds one row per swing, from the
simulator's ground-truth foot heights. The heights are sampled about
every 21\,ms of simulated time (median; 20 to 34\,ms within a trial),
so each lift-off and touchdown is known to about one sample. The
statistics use only successful trials and only swings that land on the
\emph{void field}, from when the rear feet reach the voids to the end
of the trial: \corlGaitSwingsGaitNet{} swings of
CADENCE and \corlGaitSwingsTrot{} of the trot. Since the trot failed
more often in the hardest cells, the two methods are measured on
different mixes of trials. The detector also shortens swings: the trot,
scheduled with \corlTrotSwing\,s swings on a \corlTrotPeriod\,s period,
measures a median swing of \corlGaitTrotSwingMeasured\,s, although its
median time between a leg's lift-offs, \corlGaitTrotPeriodMeasured\,s,
matches the period, and about a quarter of CADENCE's measured swings
fall below its shortest commanded swing of
\corlSwingDurationMin\,s. Hence the measured swing medians,
\corlGaitSwingMedian{} and \corlGaitSwingMedianTrot\,s, compare the two
methods under one detector and do not measure swing durations.

\noindent\textbf{Overlapping swings.}
\looseness=-1 Two swings of different legs overlap when their measured
intervals intersect. Of CADENCE's \corlGaitOverlapPairs{} overlapping
pairs, 24{,}370 (\corlGaitDiagPairPct\%) paired diagonal legs, 227 the
rear legs, 158 the two legs of one side, and 38 the front legs, whereas
all of the trot's pairs were diagonal. CADENCE's diagonal partners
lifted off a median \corlGaitLiftLagMs\,ms apart (10th to 90th
percentile, 21 to \corlGaitEventWindowMs\,ms), about one
\corlControlTickMs\,ms tick, the shortest gap that one step per tick
allows.

\noindent\textbf{Period and stepping order.}
\looseness=-1 We measure the period and the stepping order separately,
on the counted swings of each successful trial and to about one sample.
The coefficient of variation (CV) of Fig.~\ref{fig:gait}b is the
standard deviation over the mean of the times from a leg's lift-off to
its next, pooled over legs, cells, and successful trials
(\corlGaitStrideCv{} for CADENCE against \corlGaitStrideCvTrot{} for the
trot). A step rate is the number of counted lift-offs per second of
void-field time, pooled likewise (\corlGaitStepsSlow{} and
\corlGaitStepsFast{} for CADENCE at \corlBenchSpeedSlow{} and
\corlBenchSpeedFast\,m/s, and \corlGaitStepsTrot{} for the trot). The
stepping order is read from each trial's lift-offs, sorted by time and,
within a sample, by leg. A \emph{step event} groups the lift-offs within
\corlGaitEventWindowMs\,ms of its first one, the 90th percentile of the
lift-off lag between CADENCE's diagonal partners. A step event lifted
the same
set of legs as the one before it in 5{,}102 of CADENCE's 32{,}186 pairs
of consecutive events (\corlGaitEventRepeatPct\%) and in none of the
trot's 26{,}214. A leg lifted \emph{twice in a row}, that is, two
consecutive lift-offs were of the same leg, in 3{,}338 of CADENCE's
54{,}621 pairs of consecutive lift-offs (\corlGaitLegTwicePct\%) and in
none of the trot's 52{,}927. Four consecutive lift-offs covered all
four legs in 21{,}565 of CADENCE's 53{,}523 overlapping windows of four
(\corlGaitFourAllPct\%) and in 51{,}561 of the trot's 51{,}919
(\corlGaitFourAllPctTrot\%). Since a commanded swing outlasts the
event window, an event of commanded steps holds at most
\numberstringnum{\corlMaxConcurrentSwingLegs} legs under the contact
rule, so a gait that lifts each leg once per cycle in a fixed order
scores 0, 0, and 100\% on these three measures.

\end{document}